\documentclass{article}

\usepackage{microtype}
\usepackage{graphicx}
\usepackage{subfigure}
\usepackage{booktabs} 

\usepackage{hyperref}
\usepackage{enumitem}

\usepackage[accepted]{icml2025}

\usepackage{amsmath}
\usepackage{amssymb}
\usepackage{mathtools}
\usepackage{amsthm}

\usepackage[capitalize,noabbrev]{cleveref}

\theoremstyle{plain}

\theoremstyle{definition}

\theoremstyle{remark}

\usepackage[textsize=tiny]{todonotes}
\usepackage{subcaption}

\newcommand{\tsf}[1]{\text{\sf #1}}

\DeclareMathOperator*{\softmax}{\text{softmax}}

\DeclareMathOperator*{\argmin}{\text{argmin}}

\newcommand{\cX}{\mathcal{X}}

\newcommand{\E}{\mathbf{E}}
\newcommand{\R}{\mathbb{R}}
\newcommand{\rb}{\mathbb{R}}

\icmltitlerunning{Building Bridges between Regression, Clustering, and Classification}

\begin{document}

\twocolumn[
\icmltitle{Building Bridges between Regression, Clustering, and Classification}



\icmlsetsymbol{equal}{*}

\begin{icmlauthorlist}
\icmlauthor{Lawrence Stewart}{inriaens}
\icmlauthor{Francis Bach}{inriaens}
\icmlauthor{Quentin Berthet}{gdm}
\end{icmlauthorlist}

\icmlaffiliation{inriaens}{INRIA \& ENS \\ Paris, France}
\icmlaffiliation{gdm}{Google DeepMind \\ Paris, France}


\icmlcorrespondingauthor{Lawrence Stewart}{lawrence.stewart@ens.fr}

\icmlkeywords{Machine Learning, ICML}

\vskip 0.3in
]



\printAffiliationsAndNotice{\icmlEqualContribution} 

\begin{abstract}
Regression, the task of predicting a continuous scalar target $y$ based on some features $x$ is one of the most fundamental tasks in machine learning and statistics. It has been observed and theoretically analyzed that the classical approach, mean-squared error minimization, can lead to suboptimal results when training neural networks. In this work, we propose a new method to improve the training of these models on regression tasks, with continuous scalar targets. Our method is based on casting this task in a different fashion, using a target encoder, and a prediction decoder, inspired by approaches in classification and clustering. We showcase the performance of our method on a wide range of real-world datasets.
\end{abstract}

\section{Introduction}

Neural network architectures have become ubiquitous in machine learning, becoming the de facto go-to models for a wide array of tasks. This is particularly true for classification tasks, where the goal is to predict a discrete label based on observed features---e.g., in image classification \cite{krizhevsky2012imagenet, he2016deep}, language modeling \cite{sutskever2014sequence, bahdanau2015neural, vaswani2017attention}, and audio generation \cite{borsos2023audiolm, van2016wavenet}. Whilst attaining state-of-the-art results on regression problems, e.g., pose estimation, point estimation and robotics \citep{sun13deep, toshev2014deeppose, belagiannis2015robust,liu20163d}, the amount of scientific work applying neural networks to classification tasks significantly outweighs that for regression problems \citep[see, e.g.,][and references therein]{stewart2023regression}, where the objective is to predict a real-valued target $y\in \mathbb{R}^m$.

A widely observed phenomenon is that the discretization of a regression problem (sometimes referred to as ``binning'') can be beneficial for these problems. There, one transforms the real-valued labels into one-hot vectors, allowing for one to optimize the neural network's weights by minimizing the cross-entropy loss, instead of the square loss typically seen in standard regression. Real-valued predictions can be obtained from the predicted probabilities of a classification model by taking the expected value over the midpoints of the bin. Surprisingly, such discretizations can often yield better performance, despite the cross entropy loss having no notion of distance.

This behavior has been reported across a range of disciplines, e.g., computer vision \citep{zhang2016colorful, van2016pixel}, robotics \citep{rogez2017lcr,akkaya2019solving}, reinforcement learning \citep{schrittwieser2020mastering, farebrother2024stop}, biology \citep{gao2024foldtoken,picek2024geoplant}, among others \citep{lee2024binning,abe2023pathologies,ansari2024chronos}.

Understanding the cause of this pattern remains an open research problem. Analyzing the gradient dynamics of over-parametrized neural networks, \citep{globalconvergence_bach,chistikov2023learning,boursier2022gradient}, \citet{stewart2023regression} show that the implicit bias of models trained on the square loss can lead to convergence to spurious minima;  reformulating the problem to classification was observed to alleviate the under-fitting due to a change in the implicit bias. \citet{grinsztajn2022tree} observe empirically that neural networks can under-perform on regression problems due to their bias to overly-smooth solutions, as well as the lack of robustness of dense multilayer perceptron (MLP) layers to uninformative features, supporting the prior claim.

However, there are some limitations to reformulating regression problems as classification, including excessive quantization in the outputs of the model and inefficient binning of the target space, which can harm the test-time performance and also make training less efficient. In this work, we propose a generalization of these methods centered around the use of a learned target encoder-decoder pair, which allows for the end-to-end learning of the transformations that (1) generate the distributional representation of target data (i.e., the encoding), and (2) decode the distributional representation back into the target space.

These methods offer several advantages: firstly, we show that they allow for additional improvements in prediction performance over the known gain in the usual comparisons between regression and classification. One of the explanations for these improvements is that embedding the low-dimensional target space (especially when it is scalar) into an intermediate continuous space (distributions over $k$ classes) improves the training dynamics when using high-dimensional features $x \in \cX$. 



We demonstrate that these gains can be achieved even with simple architectures (a logistic model). Moreover, one can interpret the target encoder as a probabilistic latent model, which provides a smoother alternative to traditional one-hot encodings.

We also show that framing the problem in this fashion allows us to interpolate smoothly between different objectives, connecting in a natural and less binary fashion the regression and classification tasks, but also both supervised and unsupervised approaches to the target encoding. 










\paragraph{Main contributions.} In this work, we introduce a general framework for supervised regression tasks. To summarize, we make the following contributions:
\begin{itemize}[topsep=0pt,itemsep=2pt,parsep=2pt,leftmargin=10pt]

    \item We introduce a range of methods, based on the idea of target encoding into a distribution space, to improve the performance, thereby generalizing the framing of regression problems as classification.

    \item We consider in these methods a differentiable and smooth target encoding, which allows us to learn the target encoding from data, both in an unsupervised fashion from targets, and as part of a joint end-to-end loss.

    \item We showcase the improvements that our methods obtain over existing approaches over a wide range of datasets, for different data modalities, with $25 \%$ improvement in average over the least-squares baseline in regression tasks, for our fully end-to-end method.
\end{itemize}

\paragraph{Notations.} We denote by $\cX$ a general space of features, and by $\R^m$ the canonical real vector space of dimension $m$ for some positive integer $m \ge 1$, and $e_i$ the $i$-th element of its canonical basis (i.e., the one-hot vector for label $i$). For any positive integer $k$, we denote by $[k]$ the finite set $\{1, \ldots, k\}$, and by $\Delta_k \subset \R^k$ the unit simplex in dimension~$k$, of vectors with nonnegative coefficients that sum to 1. It is the convex hull of $e_1, \ldots, e_k$, and the space of discrete probabilities over $k$ elements. We denote by $H$ the entropy function from $\Delta_k$ to $\R$, defined for any $p \in \Delta_k$ by
\[
H(p) = - \sum_{i\in[k]} p_i \log(p_i)\, ,
\]
and by $\tsf{KL}$ the associated Kullback-Leibler divergence, defined for any $p, q \in \Delta_K$ by
\[
\tsf{KL}(p, q) = \sum_{i\in[k]} p_i \log\Big(\frac{p_i}{q_i}\Big)\, .
\]
We also define the $\text{softmax}$ function from $\R^k$ to $\Delta_k$, defined for $x \in \R^k$, elementwise for all $i \in [k]$ by
\[
\text{softmax}(x)_i = \frac{\exp(x_i)}{\sum_{j \in [k]} \exp(x_j)}\, .
\]
\begin{figure*}[!ht]
    \centering
    \includegraphics[width=0.85\textwidth]{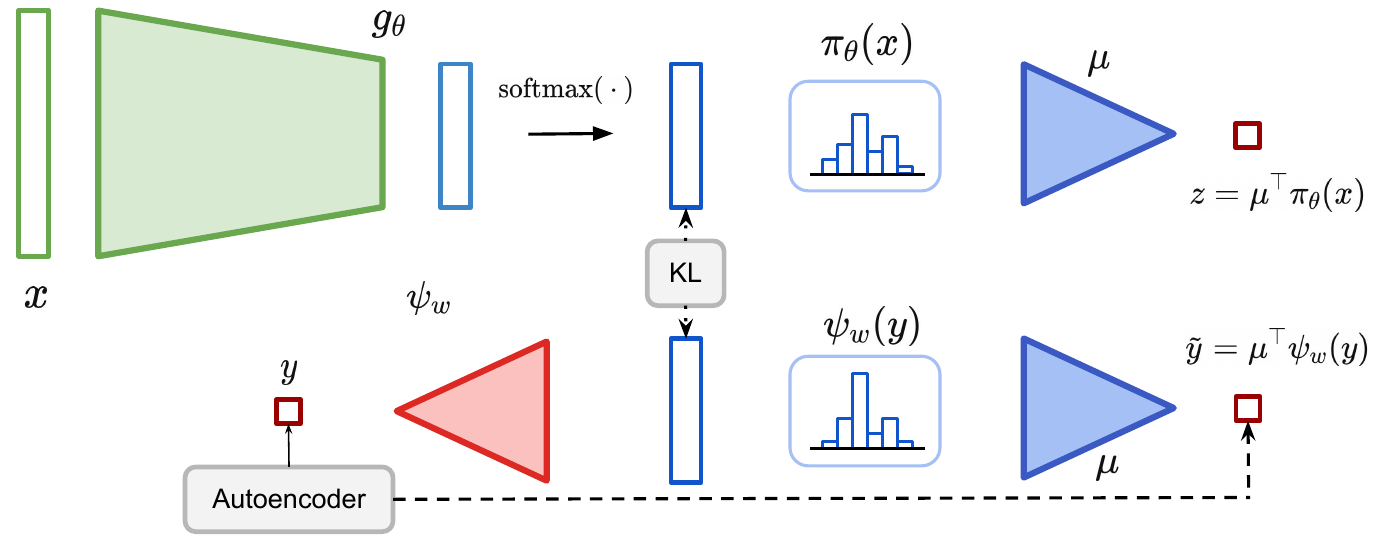}

    \caption{Framework description. Our framework is based on a \textbf{target encoder} $\psi_w$ (in red) that yields for each $y$ an encoded distribution $\psi_w(y)$ over $k$ classes. A \textbf{classification model} $\pi_\theta = \softmax(g_\theta)$ is trained with a KL objective on this distribution. A \textbf{decoder model} $\mu$ (in blue) decodes this distribution in the target space $\R^m$. The target encoder and decoder can be trained using an autoenconding loss, as well as a joint end-to-end objective (see Section~\ref{sec:methods}).}
    \label{fig:framework}
\end{figure*}

\section{Problem formulation and methods}
\label{sec:methods}

We are interested in this work in regression problems, where the aim is to infer a potentially multivariate continuous target $y \in \R^m$ based on observed features $x \in \cX$. This supervised learning problem can be tackled by using a parametrized predictor function that can be trained on a dataset of coupled examples $(x_i, y_i) \in \cX \times \R^m$, $i \in [n]$.

The several approaches to train this function that we consider follow broadly two settings and architectures, as described in Figure~\ref{fig:framework}. The most common, and most end-to-end approach is to predict directly $z = f_\eta(x) \in \R^m$, for a parametrized function (with parameter $\eta$) $$f_\eta: \cX \to \R^m,$$ and to compare it to $y$. In the approach that we propose, we consider instead several elements. The first one is a \textbf{target encoder model}, a parametrized function (with parameter~$w$) $$\psi_w:\R^m\to\Delta_k,$$ for some integer $k$. It is used to map the target $y$ to a vector of probabilities over $k$ classes. The second one is a \textbf{classification model} with logits  (parametrized by $\theta$) $$g_\theta:\cX\to\R^k,$$ used to predict a probability vector
$$\pi_\theta(x) = \softmax(g_\theta(x)) \in \Delta_k.$$ Finally, we consider a \textbf{decoder model} in the form of a linear head parametrized by a matrix $\mu \in \R^{k \times m}$, used to predict 
$$z = \mu^\top \pi_\theta(x) \in \R^m.$$
 

As discussed in further details below (see Section~\ref{sec:discussion}), this simple decoder allows for simple interpretability: each of the $k$ classes is associated to a decoder $\mu_i \in \R^m$, and the decoded prediction is an average of all the $\mu_i$, weighted by the probabilities $\pi_\theta(x)$.

\subsection{Least-squares regression}
\label{sec:least-squares}
As described above, the first classical baseline that we consider is end-to-end direct prediction of $z = f_\eta(x)$, for a parametrized function $f_\eta: \cX \to \R^m$. The parameters $\eta$ of $f_\eta$ are often trained by minimizing a loss function of the form $\ell(y,z) = L(y-z)$, with typically, $L(y-z) = \| y - z\|_2^2$. Other losses $L$ have been considered to improve robustness, such as the Huber loss \cite{huber1964robust}, or 
even nonconvex functions \citep[see, e.g.,][and references therein]{barron2019general}.

This approach consists in classical end-to-end training aiming to solve
\begin{equation}\label{eq:regression_baseline1}
    \min_{\eta} \E_{(x, y)} \big[ L (y- f_\eta(x)) \big],
\end{equation}
where $ \E_{(x, y)}$ denotes a potentially empirical expectation over features $x$ and responses $y$.

As shown in earlier work, implicit bias in regression sometimes leads to underfitting \cite{grinsztajn2022tree, stewart2023regression}. 

\subsection{Least squares with softmax layer}
\label{sec:regression-softmax}
Since several of the methods that we propose in this work (below) reframe this task using a classification model $\pi_\theta = \softmax(g_\theta)$ with outputs in $\Delta_k$ and prediction using a linear layer, with $z = \mu^\top \pi_\theta(x) \in \R^m$, we also consider the case where $f_\eta$ has this specific architecture and also compare in all our results the performance of our method with regression
\begin{eqnarray}
\label{eq:regression-softmax}
& & \min_{\theta, \mu}\! \E_{(x, y)} \big[ L (y- \mu^\top \pi_\theta(x)) \big] \\
\notag & \!\!\!\!=\!\!\! & \min_{\theta,  \mu }\! \E_{(x, y)} \big[ L (y- \mu^\top \softmax(g_\theta(x)) \big].
\end{eqnarray}
When the logit vector $g_\theta$ is a neural network output. This corresponds to adding an extra layer with $k$ neurons and a joint softmax non-linearity. The parameters $\theta$ and $\mu$ can be trained by end-to-end learning by first-order methods such as stochastic gradient descent. In our experiments (see Section~\ref{sec:expe}), this often already improves over least-squares, but not as much as using the explicit output embedding $\psi_w$. 

\begin{figure*}[t]
    \centering
    \includegraphics[width=0.85\textwidth]{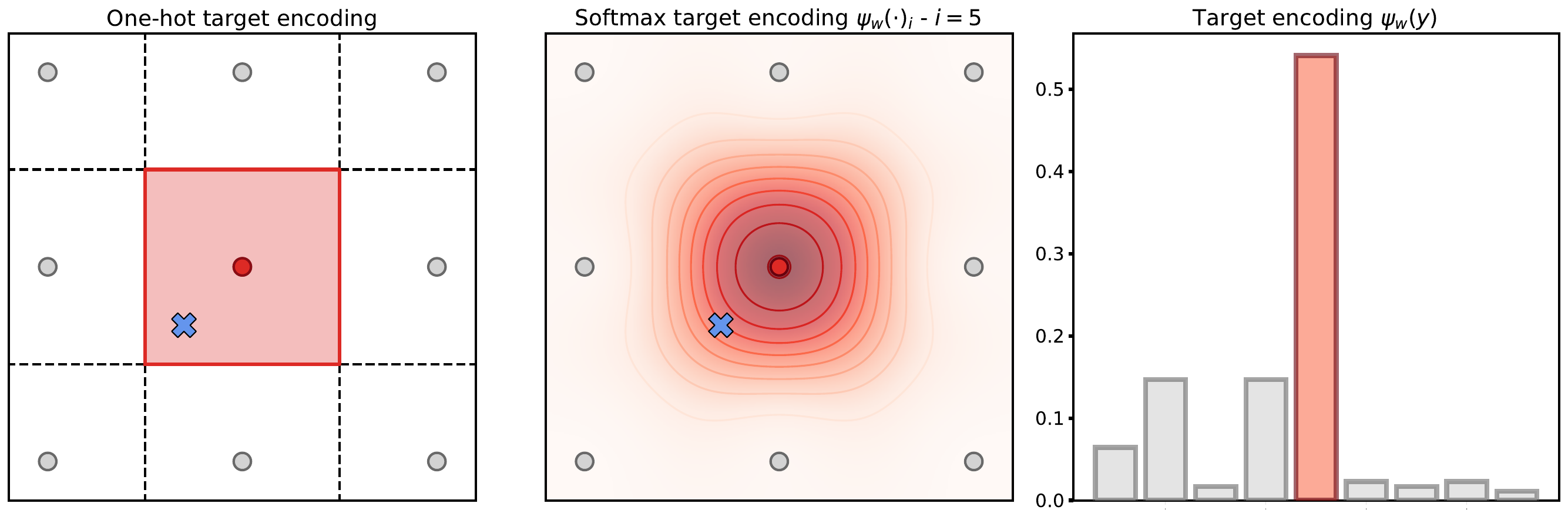}
    \hspace{1cm}
    \includegraphics[width=0.85\textwidth]{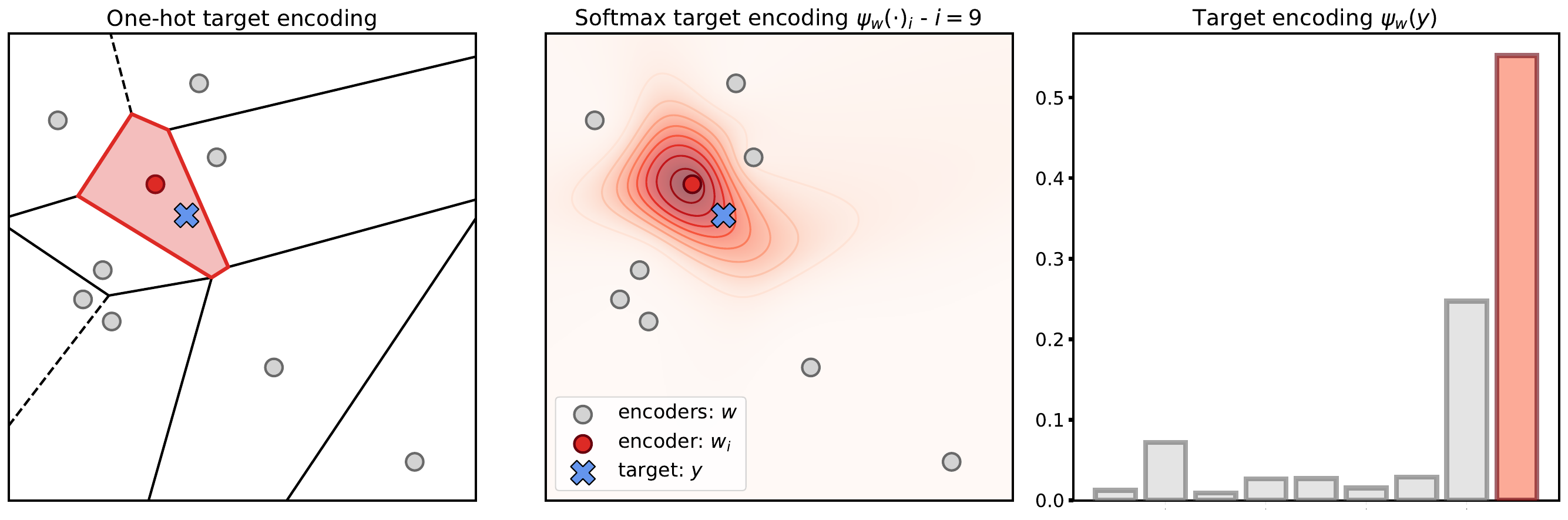}

    \caption{Embedding and binning the target space $\R^m$ (here $m=2$) into $\Delta_k$ (here $k=9$), for both a fixed grid of encoders (\textbf{Top}) and a learnt encoder (\textbf{Bottom}). For both cases we display the encoders, including an highlighted one, for a fixed $i \in [k]$ and a target $y \in \R^m$ (blue cross). We illustrate first {\em hard binning} (\textbf{Left}) where $y$ (and any $y$ in the same highlighted region) is assigned to one class (via a one-hot), and {\em soft binning} both with the contour plot of $\psi_w(\cdot)_i$ for one $i \in [k]$ (\textbf{Center}), and $\psi_w(y)$ as a distribution in $\Delta_k$ (\textbf{Right}).}
    \label{fig:embeding}
    \vspace{-0.5cm}
\end{figure*}

\subsection{Hard-binning-encoder classification} 
\label{sec:hard-binning}
An alternative existing approach is to transform the problem, reformulating it as a classification problem. This can be done by partitioning the label space $\rb^m$ (often for $m=1$), effectively by implementing with the encoder model a map $\psi_w$ from $\rb^m$ to $[k]$, represented by one-hot vectors, the extreme points of $k$-dimensional simplex $\Delta_k \subset \rb^k$.  

The main idea behind this method is to divide the target space into bins, and to identify each bin as a classification label, in order to train a classification model for prediction. This can be achieved with $k$ center points $c_1, \ldots, c_k \in \R^m$, mapping each $y$ to the label one-hot representing the nearest center---in this case $\psi_w(y) = e_i$, where $i = \argmin_{i \in [k]} \|y - c_i\|_2^2$. This approach can be interpreted as vector quantization in the target space \citep{van2017neural}.

A classification model is then trained on these newly discretized labels for the logits $g_\theta: \cX \to \R^k$ by minimizing a classification loss between the encoded target $\psi_w(y)$ and $\pi_\theta(x) = \softmax(g_\theta(x))$, such as the Kullback-Leibler divergence
\begin{equation}
\label{eq:classif-kl}
    \min_{\theta} \E_{(x, y)} \big[ \tsf{KL}(\psi_w(y) \| \pi_\theta(x))\big]\, .
\end{equation}
We note that up to a constant, this is equivalent to the more common cross-entropy loss 
\begin{eqnarray*}
    &  & \min_{\theta  } \E_{(x, y)} \big[ - \psi_w(y)^\top \log \pi_\theta(x) \big] \\
    & = &  \min_{\eta  } \E_{(x, y)} \big[ - \psi_w(y)^\top \log \softmax (g_\theta(x)) \big] \, .
\end{eqnarray*}
Indeed, these two losses only differ by a term equal to the entropy of $\psi_w(y)$. Furthermore, for this method, since the encoder maps to one-hot vectors, this entropy term is equal to 0. In a more general setting (see below) where the target encoder maps to soft vectors in the interior of the simplex, this term is either a constant (if the encoder is frozen, and we are only training the parameters $\theta$ of the classification model), or an additional term in the loss (if the two models are being trained jointly) and we state it explicitly. 

In order to use the classification model as a prediction function for $y \in \R^m$, we decode $\pi_\theta(x)$ in the target space by $z = \mu^\top \pi_\theta(x) = \mu^\top {\rm softmax}(g_\theta(x)) $ for some hand-picked decoder model $\mu \in \rb^{k \times m}$. A natural choice is to take $\mu_i = c_i$ for $i \in [k]$: when predicting the class $i$ (corresponding to targets that had $c_i$ as the nearest-center), the predicted value is $c_i \in \R^m$ (see Figure~\ref{fig:framework}).

This fixed set of encoder $\psi_w$ and linear decoder parametrized by $\mu$ has been considered for $m=1$ (e.g., by binning the $y$-space with equal size intervals or intervals with similar mass under $y$) \citep[see, e.g.][]{stewart2023regression}. This particular encoding is similar to a \textbf{clustering} approach for the target space (here done with fixed centroids). We note that other choices of one-hot encoders and decoders are also possible. 

\subsection{Soft-binning-encoder classification} 
\label{sec:soft-binning}
The first generalization that we propose in this work is to modify the classification method, by using more general target encoders that utilize the whole simplex (not only one-hot vectors). In particular, to generalize binning around $k$ centers $c_1, \ldots, c_k \in \R^m$, we consider a so-called {\em soft labels}, akin to performing a soft binning of the target space. Similarly to the previous approach, a classification model $\pi_\theta = \softmax(g_\theta)$ is then trained on these soft labels using the KL divergence as described in Equation~(\ref{eq:classif-kl}).

One way to implement this {\em soft partition} is by taking $\psi_w$ a target encoder that approximates the one-hot binning by replacing max by softmax: 
\begin{align*}
    \psi_w(y) &= \textstyle \text{softmax}\big(- \frac{\|c_1 - y\|_2^2}{2\sigma^2},\dots,- \frac{\|c_k - y\|_2^2}{2\sigma^2}\big)\\
    &=  \textstyle  \softmax \big(\frac{c_1^\top y - \frac{1}{2}\|c_1\|_2^2}{\sigma^2},\dots,\frac{c_k^\top y - \frac{1}{2}\|c_k\|_2^2}{\sigma^2}  \big)\, , \stepcounter{equation} \tag{\theequation} \label{eq:encoder_gaussian_form}
\end{align*}
for  $\sigma>0$, i.e., for all $i \in [k]$,
\begin{align*}
\psi_w(y)_i &= \frac{\exp(-\frac{1}{2\sigma^2} \|c_i - y\|_2^2)}{\sum_{j \in [k]} \exp\big(-\frac{1}{2\sigma^2} \|c_j - y\|_2^2\big)}\\
    &= \frac{\exp\big(\frac{c_i^\top y}{\sigma^2} - \frac{\|c_i\|_2^2}{2\sigma^2} \big)}{\sum_{j \in [k]}\exp\big(\frac{c_j^\top y}{\sigma^2} - \frac{\|c_j\|_2^2}{2\sigma^2} \big)}\, .\stepcounter{equation} \tag{\theequation} \label{eq:encoder_reformulated form}
\end{align*}
The two representations are mathematically equivalent (all $k$ values differ only by $\|y\|_2^2/2\sigma^2$, and the softmax function is invariant by constant shifts). The latter shows that the encoder can take a convenient form (with affine logits)
\begin{equation}\label{eq:softmax_encoder}
    \psi_w(y) = \text{softmax}(w_1^\top y+ w_2)\, ,
\end{equation}
whist the prior is connected to a classical probabilistic interpretation of softmax regression by a generative model~\citep[see, e.g.,][Section 14.2]{ltfp}, since we then have
\[
\psi_w(y)_i = \mathbf{P}(Z = i | Y=y)\, , 
\]
in a probabilistic model with a latent variable $Z \in [k]$, and isotropic Gaussian class-conditional densities with mean $c_i$ and variance $\sigma^2 I$ for the distribution of $y$ given $Z=i$. This approach, in its full generality, extends upon soft labelling methods used by, e.g.,  \citet{imani2018improving, farebrother2024stop}.

The prediction model $\pi_\theta = \softmax(g_\theta)$ is then trained by minimizing the KL divergence between $\pi_\theta(x)$ and $\psi_w(y)$, both in $\Delta_k$ as in Equation~(\ref{eq:classif-kl}).

\subsection{Pre-trained encoder} 
\label{sec:pre-trained}
The second method that we propose is a further generalization on this method, by pre-training a target encoder-decoder $(\psi_w, \mu)$, instead of hand-picking it, e.g., by minimizing an auto-encoding objective (Stage 1)
\begin{equation}
\label{eq:autoencoder}
\min_{w, \mu}\E_{y} \big[ L(y - \mu^\top  \psi_w(y)) \big]\,,
\end{equation}
and then to use this frozen target encoder to generate soft-label targets $\psi_w(y)$, to train the classification model $\pi_\theta = \softmax(g_\theta)$  as in Equation~(\ref{eq:classif-kl}) (Stage 2).

Note that the first stage can be done without access to the features $x\in \cX$, and could even be performed with synthetic data (e.g., uniform sampling on the target space if it is compact). To generalize hand-picked soft encoders, it can be chosen as a simple  model, with architecture
\[
\psi_w(y) = \text{softmax}(w_\text{lin}^\top y+ w_\text{bias})\, .
\]

Naively minimizing the auto-encoder objective in Stage 1 can afflict an implicit bias to the encoder, and yield close-to-uniform $\psi_w(y)$. To avoid this effect we can penalize the entropy, that is, minimize instead
\begin{equation}
\label{eq:loss_codec}
\min_{w, \mu} \E_y \big[ L(y-\mu^\top \psi_w(y))  - \alpha H(\psi_w(y))],
\end{equation}
with a positive parameter $\alpha>0$.


\paragraph{Initialization of encoder-decoder.}
For $m=1$, we propose initializing the decoder weights $\mu$ as a uniform spacing over the target space, where $\delta_\mu$ denotes the magnitude of the spacing. We remark that this closely resembles discretized binning \cite{stewart2023regression}. For the encoder weights, we propose setting $\sigma = \lambda_\sigma \cdot \delta_\mu$, e.g,. $\lambda_\sigma = 1$, and initializing with $c=\mu$ using the connection between Equations (\ref{eq:encoder_gaussian_form}) and~(\ref{eq:softmax_encoder}). For this initialization, the autoencoder loss $L(y - \mu^\top \psi_w(y))$ goes to $0$ for growing values of $k$, but we show experimentally that it is not necessary. For $m>1$, we suggest using a clustering algorithm such as K-means++ \citep{arthur2006k} to initialize~$\mu$. In this case $\delta_\mu$ would refer to average intra-cluster distance, and one can initialize the encoder weights in the same fashion as for $m=1$.

\begin{table*}[t]
\centering
\caption{Dataset properties. 
\label{table:datasetproperties}}
\begin{tabular}{lccccccc|c}
\hline
 & \multicolumn{7}{c}{Tabular} & \multicolumn{1}{c}{Computer Vision} \\
 \hline
 & WN & AE & BS & SC & EL & CA & DM  & RM \\
\hline
\#num. features & 7 & 33 & 6 & 79 & 16 & 21 & 6  & (3, 28, 28) \\
\#num. train points & 5,197 & 11,000 & 13,903 & 17,010 & 13,279 & 6,553 & 43,152 & 1,080 \\
\#num. val points & 650 & 1,375 & 1,738 & 2,126 & 1,660 & 819 & 5,394  & 120 \\
\#num. test points & 650 & 1,375 & 1,738 & 2,127 & 1,660 & 819 & 5,394  & 400 \\
Train batch size & 256 & 512 & 512 & 512 & 512 & 256 & 1,024  & 64 \\
\hline
\end{tabular}
\end{table*}

\subsection{End-to-end joint encoder classification}
\label{sec:end-to-end}
Our third proposed method is to combine these different objectives to jointly train the targent encoder and decoder, as well as the classification objective in Equation~(\ref{eq:classif-kl}), by minimizing the following loss, with scalar hyperparameters $\lambda_{\tsf{auto}}, \lambda_{\tsf{KL}}, \lambda_{\tsf{pred}} \ge  0$:
\begin{align*}
\min_{w, \mu, \theta} &\;
\lambda_{\tsf{auto}}\E_y \big[ L(y-\mu^\top \psi_w(y))  - \alpha H(\psi_w(y))\big] 
\\
& \hspace*{1.5cm}+\lambda_{\tsf{KL}} 
 \E_{(x, y)} \big[ \tsf{KL} \big( \psi_w(y) \| \pi_\theta(x) \big)\big] \, . \stepcounter{equation} \tag{\theequation} \label{eq:kl_and_auto}
\end{align*}

The previous approach above can be thought of as minimizing with $\lambda_{\tsf{KL}} = 0^+$---or alternatively, with $\lambda_{\tsf{KL}} = 0$ and $\lambda_{\tsf{auto}} > 0$ in Stage 1, and $\lambda_{\tsf{KL}} > 0$ and $\lambda_{\tsf{auto}} = 0$ in Stage 2. Framing it in this fashion allows for more general training of these models.

We can also add a final term that allows to stabilize the prediction loss, that is, minimize
\begin{align*}
\min_{w, \mu, \theta} &\;
\lambda_{\tsf{auto}}\E_y \big[ L(y-\mu^\top \psi_w(y))  - \alpha H(\psi_w(y))\big] 
\\
&\hspace*{1.5cm} +\lambda_{\tsf{KL}} 
 \E_{(x, y)} \big[ \tsf{KL} \big( \psi_w(y) \| \pi_\theta(x) \big)\big] \\
&\hspace*{1.5cm} + \lambda_{\tsf{pred}} \E_{(x, y)} \big[ L(y - \mu^\top  \pi_\theta(x) )\big]\, . \stepcounter{equation} \tag{\theequation} \label{eq:joint}
\end{align*}

Optimizing with different values of the loss hyperparameters $\lambda_{\tsf{auto}}, \lambda_{\tsf{KL}}, \lambda_{\tsf{pred}}$ allows us to interpolate between the different methods considered above, since it considers a linear combination of their loss objectives.


\section{Discussion}
\label{sec:discussion}

The methods that we propose are aligned with frequent observations that regression problems can be more efficiently once framed as classification problems, and in this work, we address the natural question of {\em how} they should be framed as such. Our approach to tackle this question is to use a target encoder and decoder pair, the two main advantages being that first, these models lead to {\em soft binning}, i.e., the targets are mapped not to one-hots (or labels over $k$ classes) but to whole distributions over $[k]$, and second that they are conveniently parametrized and therefore can be learnt from data, either in a two-stage, or an end-to-end fashion. As such, this work is part of a large literature on connecting discrete and continuous methods in end-to-end differentiable systems for machine learning \citep[see, e.g.][]{berthet2020learning, blondel2020fast, vlastelica2019differentiation, llinares2023deep, stewart2023differentiable}. 

Further, by smoothing over the transition between a discrete and a continuous task, the method that we propose leads to possible interpretability of the learnt codes as representations of the target data. As noted above, the decoded predictions are necessarily in the convex hull of the $\mu_i$'s, that can be interpreted as a quantization of the data. When there is an {\em a priori} natural  underlying clustering to the feature and target space, it is natural to investigate whether the learnt classes correspond to the natural ones. We observed in several experiments (see Section~\ref{sec:expe}) that the while the entropy of learnt encoded distributions $\psi_w(y) \in \Delta_k$ for targets $y$ from the data is quite low, these distributions are not typically very close to one-hots, as is more common in classification. The reason for this behavior could be connected to implicit biases and training dynamics as observed in a classification setting \cite{stewart2023regression}.

The final objective that we propose in Section~\ref{sec:end-to-end} is both strongly connected to the end-to-end paradigm of machine learning, as all objectives are jointly optimized, and going against it: naively optimizing an square loss over the same prediction $\mu^\top \pi_\theta(x)$ without considering a structured loss, with autoencoding and classification is not as performant (see Section~\ref{sec:expe}).



\begin{figure*}[t]\label{fig:rmses_datasets}
    \centering
    \includegraphics[width=0.95\textwidth]{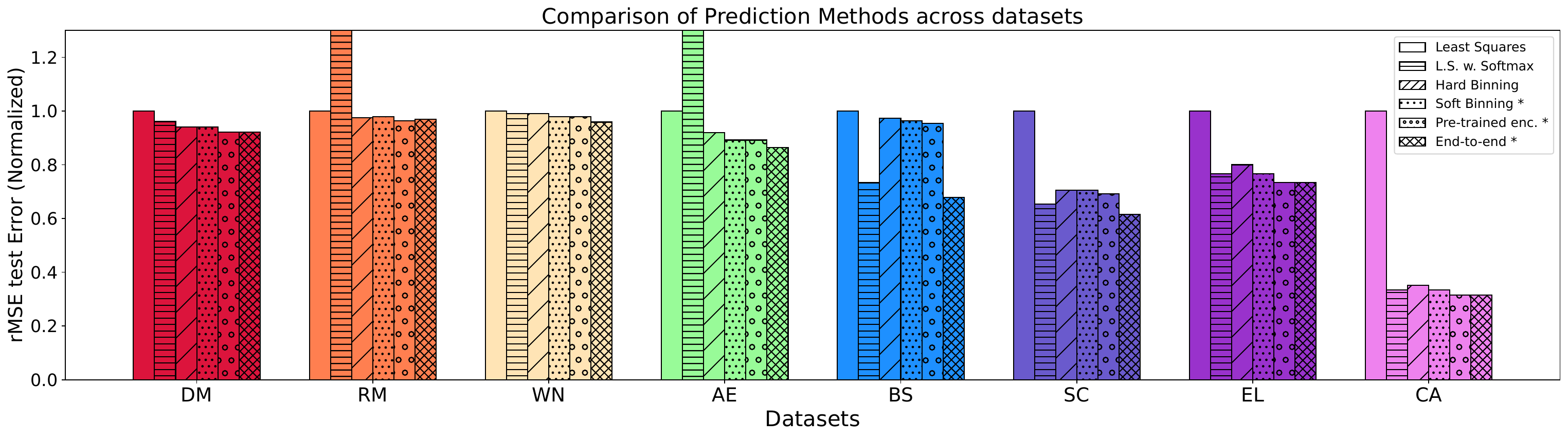}

    \caption{Experimental results across datasets. We report for all methods the test root mean squared error (rMSE), over 8 different datasets (see \textbf{Datasets} above), for the 6 methods listed in Section~\ref{sec:expe-methods}, all for $k=25$. They are displayed in each of the 8 groups from left to right. All are normalized to the error of the first baseline: least squares is set to 1.0 in each dataset, and the others proportionally. 
    }
    \label{fig:expe-barplots}
\end{figure*}

\section{Experiments}
\label{sec:expe}

\paragraph{Datasets.} We demonstrate our methodology across a diverse set of real-world regression datasets, spanning engineering, social sciences, medicine, physics, and other interdisciplinary fields, all of which are publicly available. In particular we use the following OpenML \citep{vanschoren2014openml} datasets: Ailerons (AE), Elevators (EL), Computer Activity (CA), Diamonds (DM);  the following UCI datasets: Wine Quality (WN) \citep{wine}, Bike Sharing (BS) \cite{bike_sharing}, Superconductivity (SC) \citep{superconduct}, as well as the Retina MNIST dataset (RM) from the Medical MNIST benchmark \cite{medmnistv2}. The train, validation, test split sizes and feature dimensions for each of the datasets are listed in Table \ref{table:datasetproperties}. For tabular data points, we applied min-max scaling, for images we standardize across channels, and all labels are scaled to $[0, 1]$.


\paragraph{Models.} For tabular datasets we followed the convention of prior literature \citep{gorishniy2021revisiting}, by using a multilayer perceptron (MLP), with hidden dimension 128, ReLU non-linearity, and a dropout \cite{srivastava2014dropout} of $0.3$. For image datasets we used a convolutional neural network \cite{lecun1998gradient}, using three layers of convolutions with average pooling between layers, followed by two fully-connected layers. For the convolutions, we use $(3, 3)$ kernel size, with a stride of one, and for the average pooling we use a $(2, 2)$ size with a stride of two. The two fully-connected layers have hidden dimension $256$, and use a dropout of $0.5$, with ReLU as the non-linearity. For the exact implementations of all models, data processing and training, we refer the reader to view our code repository, (implemented with PyTorch). 


\paragraph{Training.}

We trained all models using the Adam optimizer \cite{kingma2014adam} with an $\ell_2$ weight decay of $10^{-4}$ for the MLP and encoder-decoder, whilst an $\ell_2$ decay of $10^{-2}$ for the CNN. All models use a gradient clipping equal to 1. The training batch sizes for all datasets are listed in Table~\ref{table:datasetproperties}. Hyper-parameters for experiments (e.g., max learning rate, $\lambda_{\tsf{KL}}$, $\lambda_{\tsf{auto}}$, $\lambda_{\tsf{pred}}$) were selected for each model via a log-space sweep. We run repeat trials of each experiment, for which we report mean values. All experiments were ran using an NVIDIA V100 GPU.
\begin{figure}[!t] 
    \centering

    \includegraphics[width=0.35\textwidth]{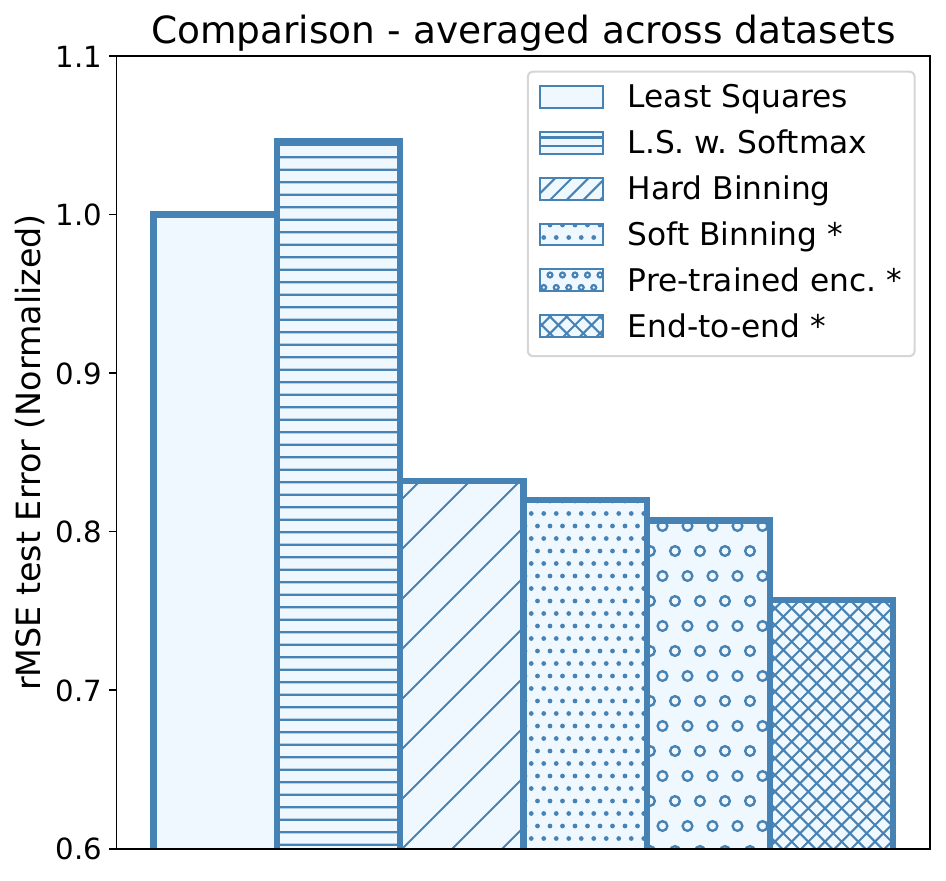} 
    \caption{Average experimental results on average over all datasets. We observe an overall hierarchy between the different methods considered. 
    }
    \label{fig:dataset_average_global}
\end{figure}

\subsection{Comparison of methods}
\label{sec:expe-methods}
For each of the datasets, we trained models by minimizing the objectives listed throughout Section \ref{sec:methods}, namely:

\begin{enumerate}
    \item \textbf{Least-squares}: The most classical, end-to-end training loss for regression as our main baseline, see Section~\ref{sec:least-squares}.
    \item \textbf{Least-squares with softmax layer}: Allows one to compare how the capacity change of the network (by effectively adding an extra layer) affects performance, see Section~\ref{sec:regression-softmax}.
    \item \textbf{Hard-binning encoder classification}: The most common practice discretization, to transform a regression problem into a classification task, as described by \citet{stewart2023regression}---see Section~\ref{sec:hard-binning}.
    \item \textbf{Soft-binning encoder classification}: We use instead a hand-picked, smooth target encoder and decoder pair, and train the model as in a classification task---see Section~\ref{sec:soft-binning}.
    \item \textbf{Pre-trained encoder classification}: We train the encoder-decoder pair (Equation \ref{eq:autoencoder}) in an unsupervised fashion on the targets, prior to classification training of the model---see Section~\ref{sec:pre-trained}.
    \item \textbf{End-to-end learning}: All terms in this task are jointly trained, as described in Equation~\eqref{eq:joint}---see Section~\ref{sec:end-to-end}. 
\end{enumerate}


\subsection{Results}
For the provided list of methodologies, we report the normalized test set root mean square error (RMSE) across all datasets, evaluating with the model weights which attained the best validation set RMSE. We evaluate all methodologies for $k\in \{5, 15, 25\}$. For $k=25$, the results across datasets are depicted in Figure \ref{fig:expe-barplots}, with the global dataset average depicted in Figure \ref{fig:dataset_average_global}. We observed the same general behavior across all values of $k$. For a table containing the full set of results, we refer the reader to Appendix~\ref{app:expe}, Table~\ref{table:all_results}.

We remark that across all datasets, reformulating regression as a classification problem via both hard-binning and soft-binning yielded improvements, with soft-binning performing globally better. This reinforces the observations of prior literature \citep{stewart2023regression,farebrother2024stop}, and secondly demonstrates the benefits of mapping targets into the interior of the simplex (our proposed initialized encoder-decoder), rather than to an extremal one-hot vector (discretized binning).

Further, we observe that training a classification model on targets generated from a trained encoder-decoder model, yielded better performance across datasets than both hand-picked soft and hard-binning. Fitting our proposed softmax encoder-decoder on the train targets is both fast and computationally light-weight, and is also promising for scenarios where the auto-encoding loss (Equation (\ref{eq:autoencoder})) at initialization can be decreased substantially (for example, in a case where $k$ is not large enough for the target distribution of $y\in \mathbb{R}^m$).

For one of our baselines, a least-squares objective for a model with softmax layer, we can see that adding the decoder's extra trainable parameters to the regression model and training with the square loss results in varied results. For some datasets (e.g., Super Conductivity), it leads to performance gains (likely due to a greater model capacity), whilst for others (and globally on average), it leads to degraded performance, even compared to the initial least-squares baseline. Our gains are therefore not due to architectural choices and the presence of a softmax layer.

Finally, it can be seen  from Figures \ref{fig:expe-barplots} \& Table \ref{table:all_results} that the proposed ``end-to-end'' objective (Equation (\ref{eq:joint})) leads to the best performance across all datasets. We stipulate this is because this approach (1) optimizes the encoder-decoder to attain a low auto-encoding error, (so decoding of classification model that has learned to predict with high accuracy the target encoding would result in a low RMSE), and (2) bridges classification and regression, with the prior potentially yielding the benefits of task reformulation, and the latter ensuring both the classification model and decoder are jointly trained by gradients coming from the regression objective.

\begin{figure}[t] 
    \centering
    \vspace{-0.2em}
    \includegraphics[width=0.45\textwidth]{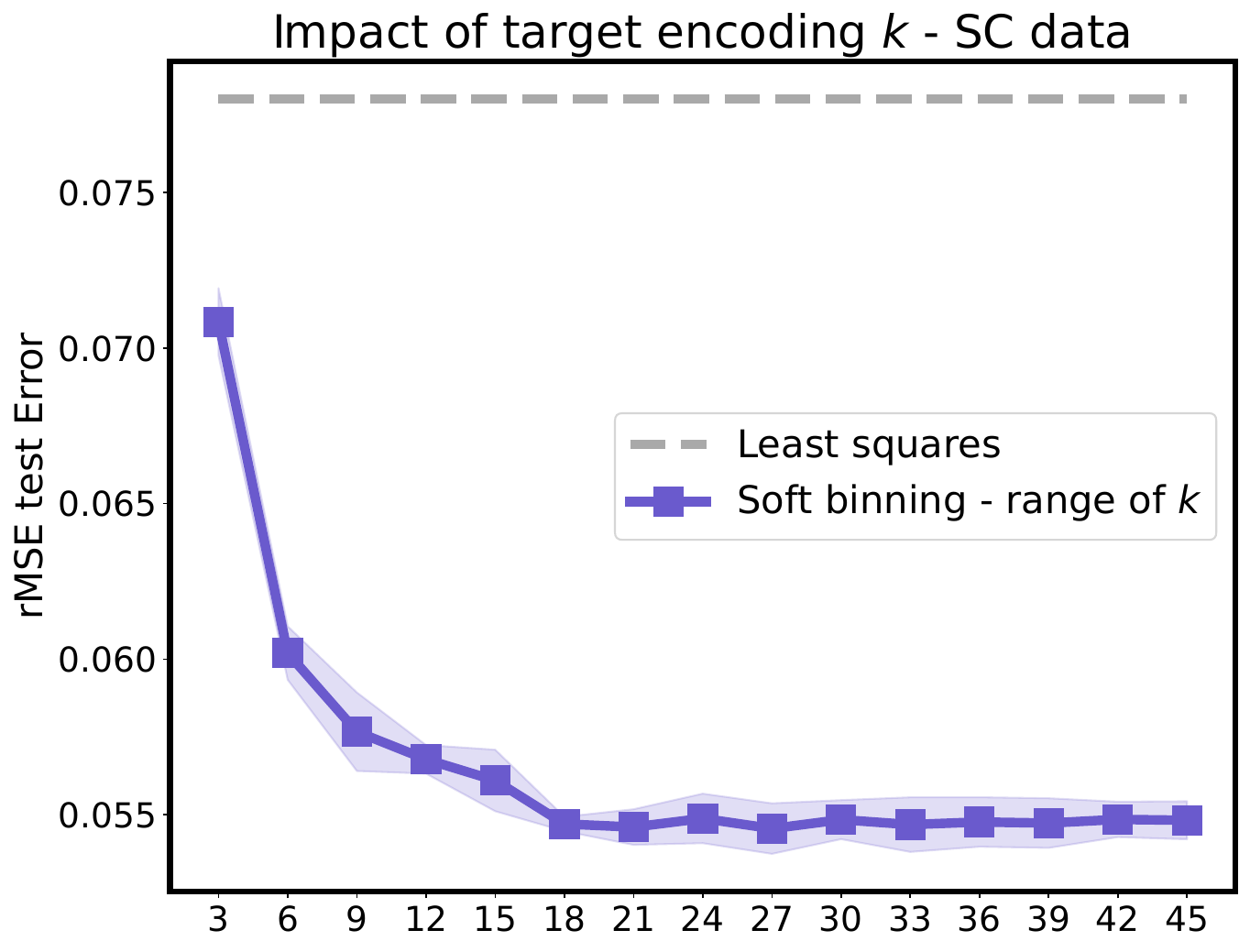} 
    \includegraphics[width=0.45\textwidth]{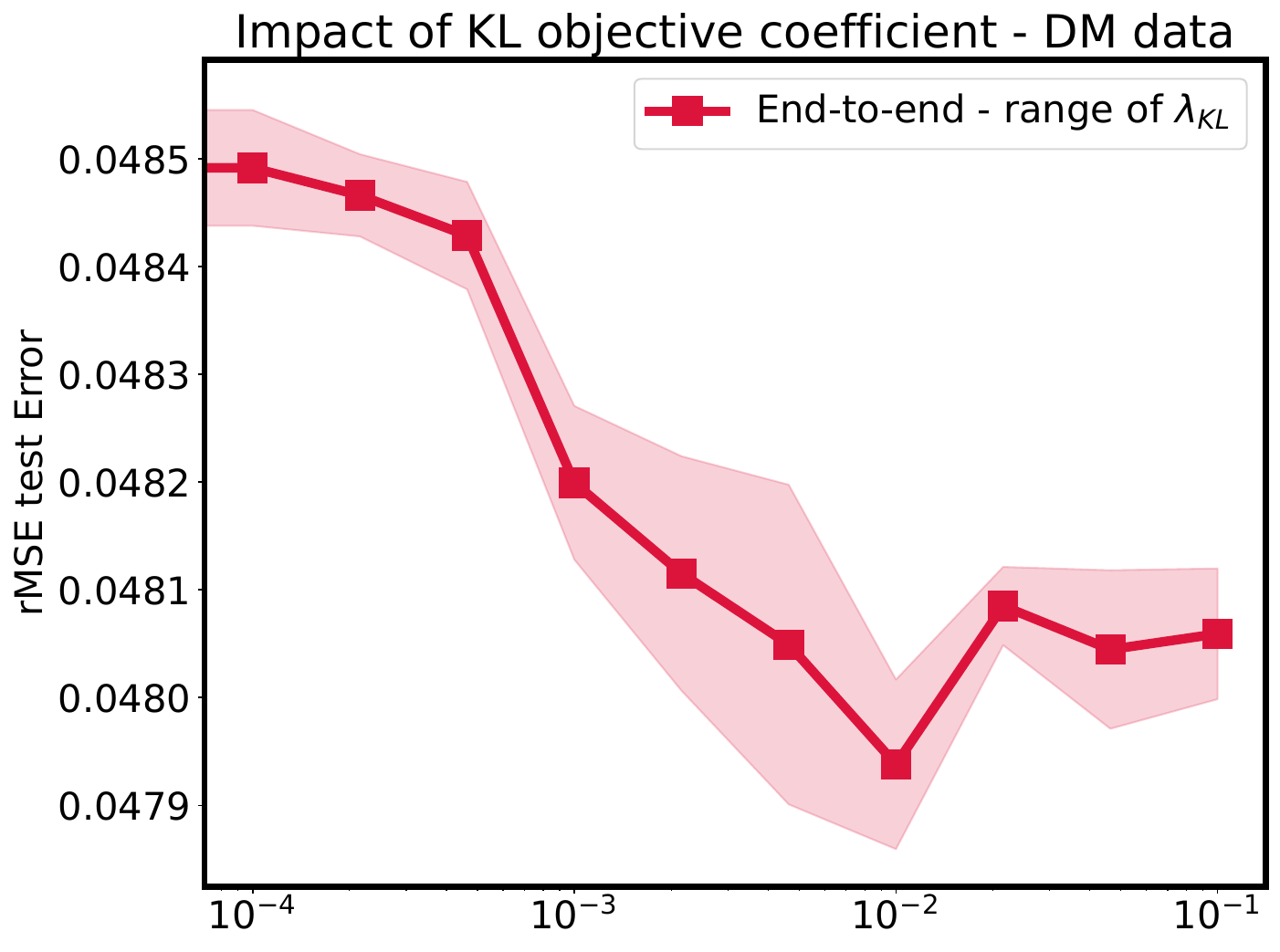} 
    \caption{Impact of different architecture and training hyperparameters on the performance of the methods. \textbf{Top}: for the soft-binning approach, the impact of $k$ for values between $3$ and $45$. \textbf{Bottom}: for the end-to-end approach, the impact of the value of $\lambda_{\tsf{KL}}$ on the final value.
    }
    \label{fig:ablation}
\end{figure}

\paragraph{Hyper-parameters.} A key hyper-parameter of our methods, as well as hard-binning is the choice of $k$, the number of classification classes (or size of the encoded distribution). For small $k$, the encoder-decoder will have less capacity to auto-encode the targets, which may hurt performance on the regression task, but larger values of $k$ yield increased optimization costs. Figure \ref{fig:ablation} depicts the relationship between $k$ and final test RMSE for soft-binning encoder classification. As $k$ increases, we can see improvements in performance, followed by a plateau with no further gains. We conclude that the choice of $k$ depends on the exact model, dataset and optimization used.

We observed empirically that when initializing the encoder-decoder weights using our proposed methodology,  the results are robust across datasets to the choice of the entropic regularization coefficient $\alpha$ in Equation (\ref{eq:loss_codec}), and taking a very small value, e.g., $10^{-6}$ suffices, (on the other hand, too large values of $\alpha$ will negatively affect the auto-encoding loss listed in Equation (\ref{eq:autoencoder})).

For end-to-end training, we observed that it is important to find a good balance between the classification and regression loss terms, via the choice of $\lambda_{\tsf{KL}}$ and $\lambda_{\tsf{pred}}$. 
Whilst for some select datasets and values of $k$, we observed that training with only the $\tsf{KL}$ objective on fitted encodings produced similar performance, we overall found that there were no single values of $\lambda_{\tsf{KL}}$, $\lambda_{\tsf{pred}}$ that were optimal for all datasets. In general we set $\lambda_{\tsf{pred}} = 1$ and performed a sweep to find $\lambda_{\tsf{KL}}$, the best values for each dataset being listed in Table \ref{table:kl_sweep} 
within Appendix \ref{app:expe}. Figure \ref{fig:ablation} depicts impact of this parameter for the DM dataset, and highlights how the combination of the regression and classification loss can lead to better results than just one of the losses. We remark that overall, the dependency of the final results on these hyper-parameters was low, indicating a robustness to these choices.


\paragraph{Conclusion.} For regression problems we have proposed  introducing a light-weight target encoder-decoder, trained jointly (or frozen) with a classification model using a loss (Equation \ref{eq:joint}) that balances regression, classification and auto-encoding of the targets. We empirically explore the effect of each of our proposed generalizations (Section \ref{sec:methods}), as well as ablating hyper-parameter choices. Notably, our end-to-end method consistently outperforming the prior regression and classification baselines \cite{stewart2023regression}, across a wide range of real-world datasets.

\newpage
 \section*{Acknowledgments}
The authors would like to thank Tianlin Liu, Michaël Sander and David Holzmuller for their help and feedback on preliminary versions of this work. The French government partly funded this work under the management
of Agence Nationale de la Recherche as part of the “France 2030” program, reference
ANR-23-IACL-0008 (PR[AI]RIE-PSAI).
 
 \section*{Impact Statement}

This paper presents work whose goal is to advance the field of 
Machine Learning, by proposing improvements to regression tasks in supervised learning. There are many potential societal consequences 
of our work, none which we feel must be specifically highlighted here.

\bibliography{main}
\bibliographystyle{icml2025}

\clearpage
\appendix
\onecolumn


\section{Additional experimental results}
\label{app:expe}
We provide further experimental details

\begin{table}[!h]

\centering
\caption{Test set RMSE (averaged over random seeds) for all methodologies across datasets. \label{table:all_results}}
\begin{tabular}{ccccccccc}
\toprule
 & WN & AE & BS & SC & EL & CA & DM  & RM   \\
\midrule
Least Squares & 0.097 & 0.037 & 0.109 & 0.078 & 0.03 & 0.057 & 
0.051  & 0.195 \\
\midrule
Least Squares w. Softmax $k=5$  & 0.097 & 0.089 & 0.090 & 0.055 & 0.024 & 0.022 & 0.049  & 0.200 \\
Least Squares w. Softmax  $k=15$  & 0.095 & 0.089 & 0.079 & 0.051 & 0.023 & 0.019 & 0.049  & 0.298 \\
Least Squares w. Softmax  $k=25$  & 0.095 & 0.089 & 0.080 & 0.051 & 0.023 & 0.019 & 0.049  & 0.298 \\

\midrule 
Hard Binning Classification $k=5$ & 0.095 & 0.042 & 0.106 & 0.057 & 0.042 & 0.041 & 
0.050  & 0.210 \\
Hard Binning Classification $k=15$ & 0.095 &0.034 & 0.106 & 0.055 & 0.025 & 0.020 & 
0.048  & 0.195 \\
Hard Binning Classification  $k=25$ & 0.096 &0.034 & 0.106 & 0.055 & 0.024 & 0.020 & 
0.048  & 0.195 \\
\midrule 
Soft Binning Classification $\lambda_\sigma = 1$, $k=5$  & 0.096 & 0.043 & 0.132 & 0.096 & 0.060 & 0.067 & 0.055  & 0.210 \\
Soft Binning Classification $\lambda_\sigma = 1$, $k=15$  & 0.095 & 0.033 & 0.108 & 0.058 & 0.025 & 0.019 & 0.048  & 0.193 \\
Soft Binning Classification $\lambda_\sigma = 1$, $k=25$  & 0.096 & 0.033 & 0.106 & 0.057 & 0.024 & 0.019 & 0.048  & 0.190\\
\midrule
Soft Binning Classification $\lambda_\sigma = 0.5$, $k=5$  & 0.095 & 0.033 & 0.107 & 0.062 & 0.027 & 0.019 & 0.049  & 0.190 \\
Soft Binning Classification $\lambda_\sigma = 0.5$, $k=15$  & 0.095 & 0.033 & 0.105 & 0.056 & 0.024 & 0.019 &0.048  & 0.190 \\
Soft Binning Classification $\lambda_\sigma = 0.5$, $k=25$  & 0.095 & 0.033 & 0.105 & 0.055 & 0.023 & 0.019 & 0.048  & 0.191 \\
\midrule
Trained Encoder Classification $k=5$  & 0.095 & 0.033 & 0.107 & 0.058 & 0.025 & 0.019 & 0.049  & 0.189 \\
Trained Encoder Classification $k=15$  & 0.095 & 0.033 & 0.105 & 0.055 & 0.023 & 0.018 & 0.048  & 0.188 \\
Trained Encoder Classification $k=25$  & 0.095 & 0.033 & 0.104 & 0.054 & 0.022 & 0.018 & 0.047  & 0.188 \\
\midrule
End to End $k=5$ & 0.095 & 0.033  & 0.091 & 0.054 & 0.024 & 0.018 & 0.048  & 0.189 \\
End to End $k=15$ & 0.093 & 0.033   & 0.080 & 0.049 & 0.022 & 0.018 & 0.048  &  0.189 \\
End to End $k=25$ & 0.093  &  0.032  & 0.074 & 0.048 & 0.022 &  0.018 & 0.047  & 0.189 \\
\bottomrule
\end{tabular}
\end{table}

\begin{table}[!h]

\centering
\caption{Optimal sweep $\lambda_{\tsf{KL}}$ for datasets, with fixed $\lambda_{pred}=1$.\label{table:kl_sweep}}
\begin{tabular}{ccccccccc}
\toprule
 & WN & AE & BS & SC & EL & CA & DM  & RM   \\
 \midrule
 $k=5$ & 0.068 & 0.068  & 0.068 & 0.068 & 0.068 & 0.068 & 0.068  & 0.147 \\
 $k=15$ & 0.068 & 0.068  & 0.068 & 0.147 & 0.068 & 0.068 & 0.068  & 3.163 \\
 $k=25$ & 0.316 & 0.147  & 0.068 &  0.068 & 0.147 & 0.068 & 0.068  & 0.147 \\
\bottomrule
\end{tabular}
\end{table}

\section{Additional figures}

\begin{figure*}[!ht]
    \centering
    \includegraphics[width=\textwidth]{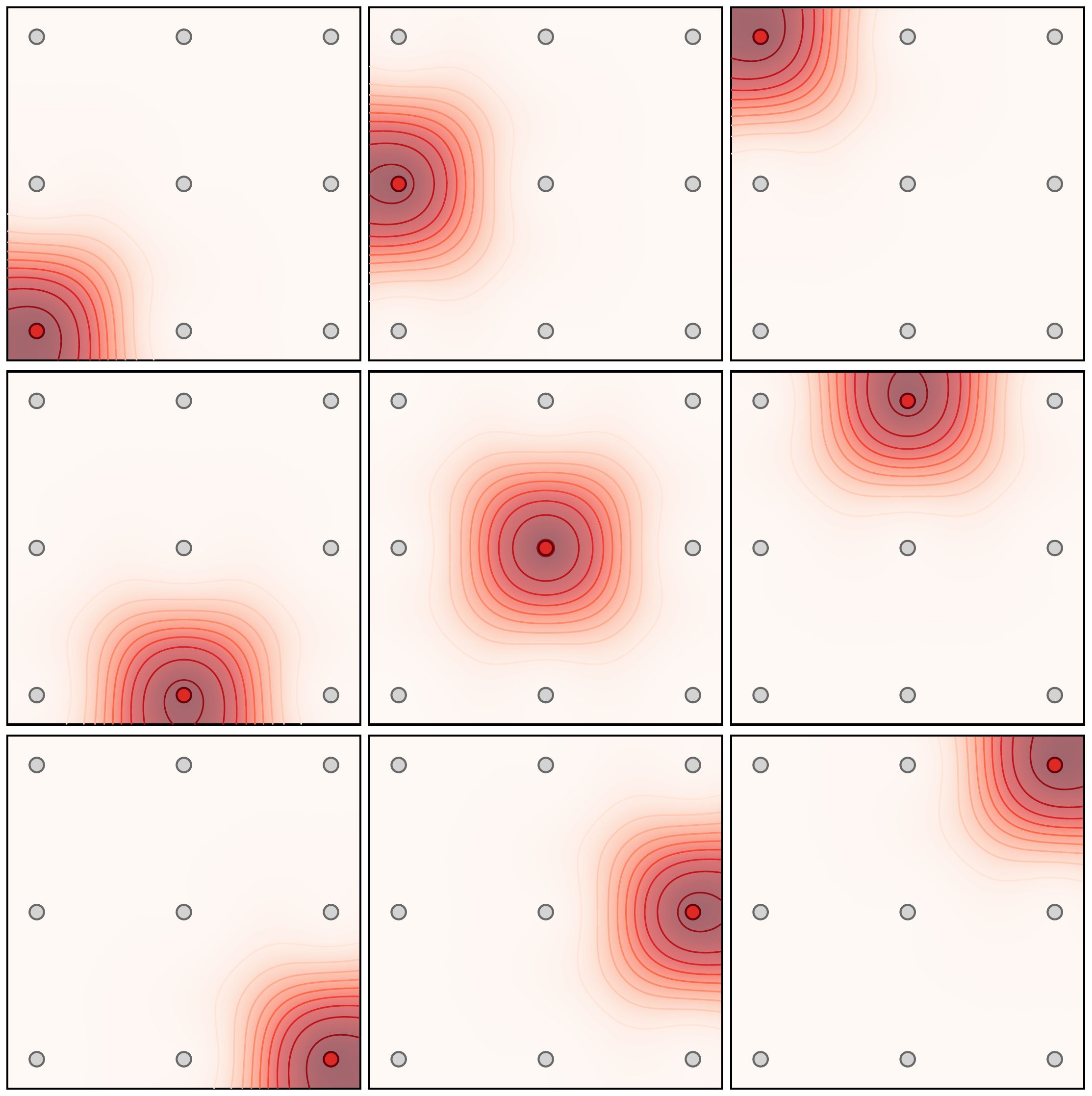}

    \caption{Embedding the target space $\R^m$ (here $m=2$), representation of all the coefficients of the target encoding}
    \label{fig:all_psi_grid}
\end{figure*}

\begin{figure*}[!ht]
    \centering
    \includegraphics[width=\textwidth]{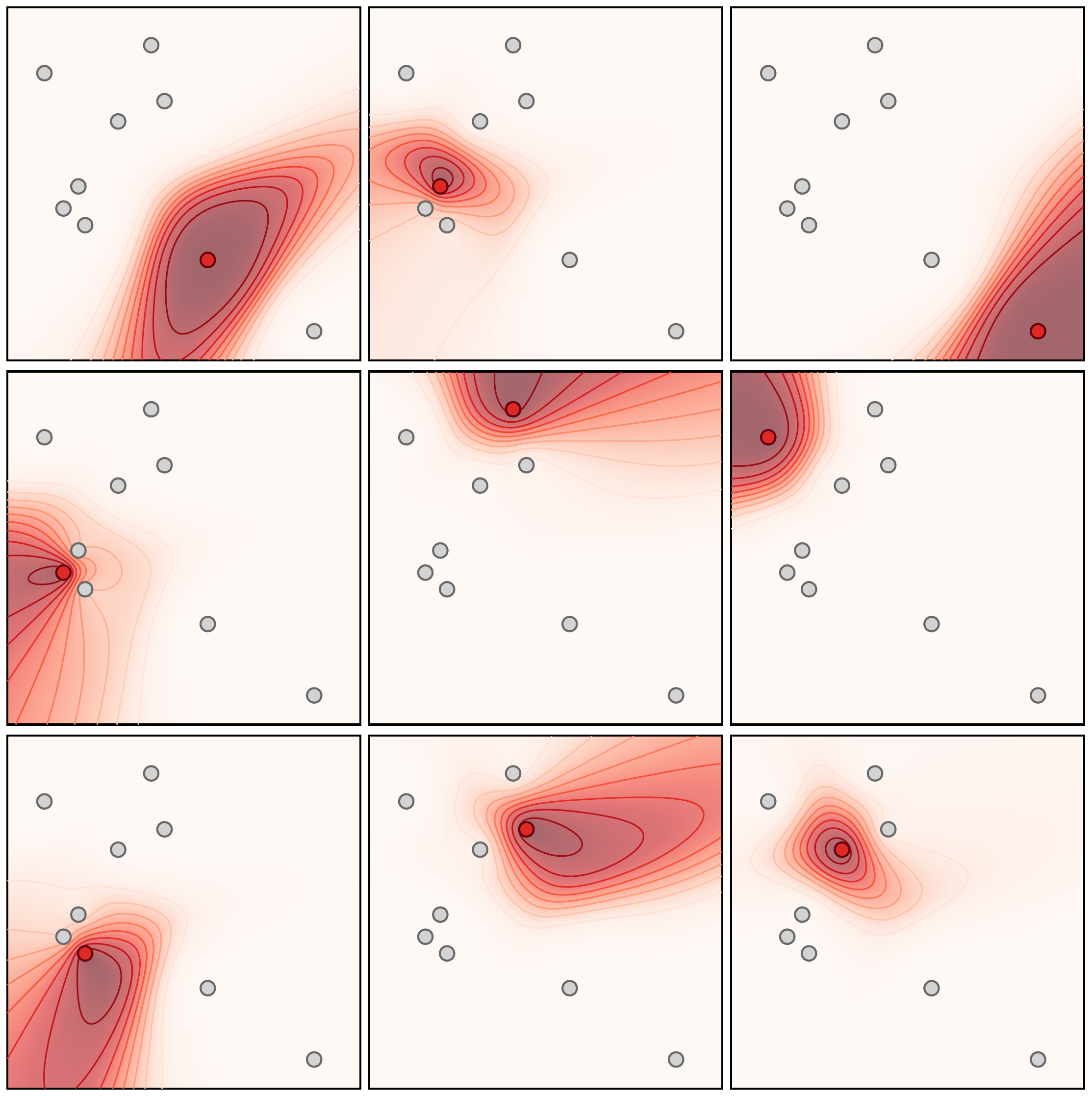}

    \caption{Embedding the target space $\R^m$ (here $m=2$), representation of all the coefficients of the target encoding}
    \label{fig:all_psi_learnt}
\end{figure*}

We provide more detailed illustration of the target encoding functions $\psi_w(\cdot)_i$ over $\R^2$ for $i \in \{1,\ldots, 9\}$ from Figure~\ref{fig:embeding}, in Figures~\ref{fig:all_psi_grid} \& \ref{fig:all_psi_learnt} below

\end{document}